\documentclass[]{fairmeta}

\usepackage{amsmath}
\usepackage{amssymb}
\usepackage[shortlabels]{enumitem}

\microtypesetup{expansion=false}

\title{Beyond Semantic IDs: Atomic Creator Tokens for Large-Scale Generative Recommendation}

\author[1]{Zhuosheng Liu}
\author[1]{Yihan Wu}
\author[1]{Ziyun Xu}
\author[1]{Xiaoxuan Liu}
\author[1]{Longfei Yun}
\author[1]{Kevin Han}
\author[1]{Mingze Gao}
\author[1]{Pengfei Wang}
\author[1]{Junfeng Pan}
\author[1]{Wenjie Fu}
\author[1,\dagger]{Haoran Liu}

\affiliation[1]{Meta Platforms, Menlo Park, CA, USA}

\contribution[\dagger]{Corresponding author}

\abstract{Recommending creators that a user does not yet follow is a cold-start problem: from hundreds of
thousands of popular creators and only a short trail of prior engagements, the system must predict
the next creator worth surfacing. The prevailing generative recommendation recipe tokenizes each
creator into a \emph{Semantic ID} (SID) by quantizing a frozen embedding and trains a language model
to generate SIDs. We show that this SID layer, not the language model, is the binding constraint.
The creator-to-SID map is many-to-one, so colliding creators must be resolved by a popularity
tie-break that caps creator-level recall and favors already-popular creators, and the frozen
quantizer stops the training gradient from reaching the creator representation. We propose
\textbf{Atomic Creator Token (ACT)}, which gives every creator its own vocabulary token trained
end-to-end. The map is bijective, so collisions are impossible and inference is a direct top-$k$
over creator logits. In a controlled comparison on production data in which the token layer is the
only variable, ACT improves creator-level Recall@15 by 19.8 points ($1.84\times$) over a matched
SID baseline, still leads when the SID codebook is enlarged $64\times$, and serves at
${\sim}2.7\times$ lower per-user latency. The advantage carries over to the public KuaiRand
dataset, where ACT is the only method that beats a global-popularity baseline, provided each
creator token sees enough training examples. In a live A/B test, ACT is directionally positive on
every reported metric against both the SID-enhanced and the classic production rankers.
}

\date{\today}
\correspondence{Haoran Liu at \email{ryan0814@meta.com}}

\begin{document}

\maketitle

%% sections/introduction.tex
\section{Introduction}
\label{sec:intro}

Recommending creators that a user does not already follow, i.e., \emph{unconnected creator
recommendation}, is a cold-start problem at ecosystem scale. Out of a multi-million catalog, the
system must pick from hundreds of thousands of popular creators the few a user will enjoy, with no
follower connection and only a short, sparse trail of prior engagements (opens, dwells, follows) on
previously suggested creators. Because this signal is so limited, how the model represents
creators largely determines what it can recommend. We focus on the \emph{head} of the catalog, the
creators above a follower threshold that a platform can realistically promote and where engagement
concentrates: as the head vocabulary grows it covers a rapidly rising share of engagement, so a
larger head vocabulary delivers more \emph{effective} recall (\Cref{sec:exp:coverage}).

The dominant generative recipe, exemplified by OneRec~\citep{onerec2025} and the Transformer Index
for Generative Recommenders (TIGER)~\citep{rajput2023recommender}, recasts recommendation as
sequence generation over \emph{Semantic IDs} (SIDs). A pretrained graph neural network (GNN) or
two-tower model produces a frozen creator embedding, which residual
quantization~\citep{douze2025faiss} compresses into a short sequence of codebook indices; a
language model is then fine-tuned to generate the next SID, and the generated SIDs are decoded back
to creators through a codebook join and a popularity rerank.

We argue that this SID layer, not the language model, is the binding constraint, for two
compounding reasons. First, the creator-to-SID map is \emph{surjective}: the used code space is far
smaller than the catalog, so many creators share one SID. Even a model that predicts the right SID
must still guess which of its creators the user wants, and the popularity tie-break that resolves
this guess caps creator-level recall and favors already-famous creators, the opposite of what
surfacing fresh creators requires. Second, the quantizer is \emph{frozen and non-differentiable}:
the next-token gradient never reaches the creator representation, so the tokenizer and the language
model optimize mismatched objectives. The same design also forces a codebook rebuild for every new
creator and a multi-step decode-and-join at serving time.

\begin{figure}[t]
  \centering
  \includegraphics[width=\textwidth]{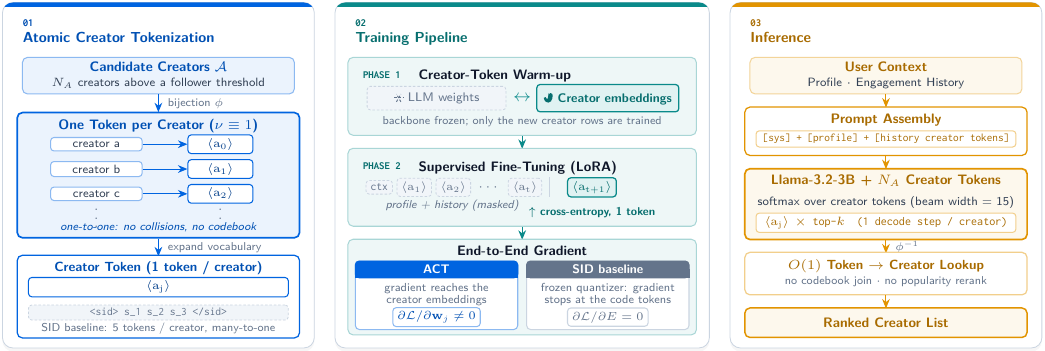}
  \caption{Overview of ACT. (01) Each candidate creator gets one dedicated vocabulary token
  through a bijection $\phi$, instead of a many-to-one Semantic ID. (02) The creator embeddings are
  warmed up with the backbone frozen and then fine-tuned end-to-end with Low-Rank Adaptation
  (LoRA), so the gradient reaches them. (03) At inference, a single decode step per creator yields
  a top-$k$ over creator tokens, decoded by an $O(1)$ lookup without a codebook join or popularity
  rerank.}
  \label{fig:overview}
\end{figure}

We therefore propose \textbf{Atomic Creator Token (ACT)}, which removes the SID layer entirely
(\Cref{fig:overview}). ACT expands the language model's vocabulary with one dedicated token per
candidate creator, up to $30Y$ tokens in our study, and trains the model end-to-end to generate the
next creator token. Because the creator--token map is \emph{bijective}, collisions are impossible
and inference reduces to a direct top-$k$ over creator logits with no codebook join or popularity
rerank. Because the creator embeddings are model parameters, the next-token gradient trains them
directly, which also makes on-policy alignment from engagement feasible. The added capacity is
modest: a $10Y\times3072$ embedding table in bf16 adds ${\approx}0.6$\,GB.

ACT is not a drop-in replacement for arbitrarily large catalogs. One token per creator needs enough
engagement examples per token to learn a useful embedding, and the output softmax grows with the
vocabulary, so we position ACT as the head component of a \emph{routed} recommender: ACT covers
every above-threshold creator, whereas a frozen SID codebook reaches only the $86.3\%$ with a GNN
embedding, and conventional retrieval serves the sparse long tail. Our public-data study makes this
premise explicit (\Cref{sec:exp:public}). Our contributions are:
\begin{enumerate}[leftmargin=*, nosep]
  \item a diagnosis of the SID bottleneck in cold-start creator recommendation: a surjective
  creator-to-SID map imposes a collision-limited recall ceiling, and a frozen quantizer blocks
  end-to-end representation learning;
  \item \textbf{ACT}, a generative recommender with a bijective per-creator vocabulary trained
  end-to-end, which is collision-free and join-free at inference and fits in a single-host memory
  budget at $30Y$ creators;
  \item a controlled evaluation on production data in which the token layer is the only variable:
  ACT improves creator-level Recall@15 by $+19.8$ points ($1.84\times$) over a matched SID
  baseline at a $10Y$-creator vocabulary, still leads when the SID codebook is enlarged to
  $(4X)^3$ at $30Y$ creators, and is cheaper to serve;
  \item a replication on the public KuaiRand dataset, together with an analysis of when ACT
  helps: it wins once each creator token has enough training examples, and a live A/B test that
  is directionally positive on every reported metric.
\end{enumerate}

%% sections/method.tex
\section{Method}
\label{sec:method}

We first define the task and the SID baseline that ACT replaces (\Cref{sec:method:sid}), then
describe ACT's tokenization and objective (\Cref{sec:method:act}) and its training pipeline
(\Cref{sec:method:train}).

\subsection{Task and Semantic-ID Generation}
\label{sec:method:sid}

\paragraph{Task.} Let $\mathcal{U}$ be the set of users and $\mathcal{A}$ the set of candidate
creators. Each user $u\in\mathcal{U}$ has profile features $x_u$ (country, language, gender, age)
and a chronological engagement history $H^u_t=\big((a_1,\tau_1),\dots,(a_t,\tau_t)\big)$, where
$a_i\in\mathcal{A}$ is the $i$-th engaged creator and $\tau_i$ its engagement type (open, like,
share, follow, or long dwell). An engagement-value function $v(\tau)\ge0$ scores these types,
increasing from a brief open to a follow; we use it to weight NDCG and to shape rewards. The task
is next-creator prediction: a model $P_\theta(a_{t+1}\mid H^u_t,x_u)$ with parameters $\theta$
returns the $K$ highest-scoring creators $\widehat{\mathcal{A}}^K_i$ for each test instance $i$, and
we report
\begin{equation}
\mathrm{Recall}@K=\frac{1}{N}\sum_{i=1}^{N}\mathbb{1}\!\left[a^\star_i\in\widehat{\mathcal{A}}^K_i\right],
\end{equation}
where $N$ is the number of test instances, $a^\star_i$ is the ground-truth next creator of instance
$i=1,\dots,N$, and $\mathbb{1}[\cdot]$ is the indicator function.

\paragraph{Semantic IDs.} The SID baseline maps each creator to an embedding with a pretrained,
frozen encoder $E:\mathcal{A}\to\mathbb{R}^{d_E}$ (a GNN) and quantizes it with $L$ codebooks, each
holding $M$ centroids $\{\mathbf{e}_{\ell,k}\}_{k=1}^{M}$, by greedily quantizing successive
residuals:
\begin{equation}
r^{(0)}=E(a),\qquad
s_\ell(a)=\arg\min_{k\in[M]}\big\|r^{(\ell-1)}-\mathbf{e}_{\ell,k}\big\|_2,\qquad
r^{(\ell)}=r^{(\ell-1)}-\mathbf{e}_{\ell,s_\ell(a)},
\end{equation}
where $d_E$ is the embedding dimension, $r^{(\ell)}$ is the residual left after level
$\ell=1,\dots,L$, $\mathbf{e}_{\ell,k}$ is the $k$-th centroid of codebook $\ell$, and
$[M]=\{1,\dots,M\}$. The SID of creator $a$ is $s(a)=(s_1(a),\dots,s_L(a))\in[M]^L$, and
\begin{equation}
\nu(s)=\big|\{a\in\mathcal{A}:s(a)=s\}\big|
\end{equation}
is the \emph{collision count} of code $s$, the number of creators that share it. Because the
reachable code space is effectively far smaller than $|\mathcal{A}|$, the map $s$ is surjective and
$\nu(s)>1$ for many codes.

\paragraph{Generation and inference.} A language model generates the $L$ code tokens of the next
SID autoregressively, $P_\theta(s\mid H,x)=\prod_{\ell=1}^{L}P_\theta(s_\ell\mid s_{<\ell},H,x)$,
and is trained with supervised fine-tuning (SFT). Its gradient stops at the SID-token embeddings, so
neither $E$ nor the codebooks adapt to the recommendation objective. At inference, beam search
produces candidate SIDs, each SID is expanded to its $\nu(s)$ creators by a codebook join, and a
popularity score breaks the ties before the top-$K$ creators are returned.

\subsection{Atomic Creator Tokens}
\label{sec:method:act}

ACT keeps this generative setup but removes the SID layer. It assigns each creator a unique token
through a bijection $\phi:\mathcal{A}\to\{0,1,\dots,N_A-1\}$, where $N_A=|\mathcal{A}|$, and expands
the base vocabulary $\mathcal{V}$ of the language model with the creator tokens
$\{\langle a_j\rangle\}_{j=0}^{N_A-1}$. Whereas the SID map has $\nu(s)\ge1$, ACT has $\nu\equiv1$:
no two creators share a token, and decoding a token back to a creator is an $O(1)$ lookup
$\phi^{-1}$ instead of a one-to-many codebook join.

\paragraph{Model.} We use a Llama-3.2-3B backbone~\citep{grattafiori2024llama} with hidden size
$d=3072$. Each creator token adds one ordinary vocabulary embedding
$\mathbf{w}_{\phi(a)}\in\mathbb{R}^{d}$, no different in kind from the model's word-piece
embeddings. A context $z=(x_u,H^u_t)$ is serialized into a prompt, with the profile as text and the
history as the sequence of the engaged creators' tokens, and the backbone $f_\theta$ produces the
hidden state $\mathbf{h}=f_\theta(z)\in\mathbb{R}^{d}$ at the position that predicts the next
creator. With decoding restricted to creator tokens, the next-creator distribution is the usual
next-token softmax over $\mathcal{A}$:
\begin{equation}
P_\theta(a\mid z)=\frac{\exp\!\big(\mathbf{w}_{\phi(a)}^\top\mathbf{h}\big)}{\sum_{a'\in\mathcal{A}}\exp\!\big(\mathbf{w}_{\phi(a')}^\top\mathbf{h}\big)},
\label{eq:act-softmax}
\end{equation}
where $a'$ ranges over all candidate creators. A creator is thus emitted as a single token rather
than an $L$-token SID.

\paragraph{Objective.} We train with the next-token cross-entropy on the target creator,
\begin{equation}
\mathcal{L}_{\text{SFT}}=-\mathbb{E}_{(z,a^\star)}\big[\log P_\theta(a^\star\mid z)\big],
\end{equation}
where the expectation $\mathbb{E}$ is over training pairs of a context $z$ and its ground-truth next
creator $a^\star$. There is no separate creator encoder: assigning each creator a vocabulary token
is the entire representational change. Because the creator embeddings live inside the model, the
loss updates each of them directly,
\begin{equation}
\nabla_{\mathbf{w}_j}\mathcal{L}_{\text{SFT}}=\big(p_j-\mathbb{1}[\,j=\phi(a^\star)\,]\big)\,\mathbf{h},
\qquad p_j=P_\theta\big(\phi^{-1}(j)\mid z\big),
\end{equation}
where $j=0,\dots,N_A-1$ indexes the creator tokens and $p_j$ is the predicted probability of
creator $\phi^{-1}(j)$. The SID pipeline, in contrast, has a zero gradient with respect to its
frozen encoder $E$.

\paragraph{Why direct tokens dominate.} An SID is a deterministic, lossy function of its creator,
so by the data processing inequality an SID history carries no more information about the next
SID than a creator history carries about the next creator. The best next-token loss ACT can reach
is therefore no worse than that of any SID scheme, for any vocabulary size.

\paragraph{Contrastive auxiliary loss (optional).} To sharpen the discriminability of creator
tokens, an in-batch contrastive term can be added over a batch $\mathcal{B}$ of context--creator
pairs, treating the other creators in the batch as negatives:
\begin{equation}
\mathcal{L}_{\text{con}}=-\frac{1}{|\mathcal{B}|}\sum_{i\in\mathcal{B}}\log
\frac{\exp\!\big(\mathbf{h}_i^\top\mathbf{w}_{\phi(a_i^\star)}/\kappa\big)}{\sum_{j\in\mathcal{B}}\exp\!\big(\mathbf{h}_i^\top\mathbf{w}_{\phi(a_j^\star)}/\kappa\big)},
\end{equation}
where $\mathbf{h}_i$ and $a_i^\star$ are the hidden state and target creator of pair $i$, the
index $j$ runs over the pairs in $\mathcal{B}$, and $\kappa>0$ is a temperature; training then
minimizes $\mathcal{L}_{\text{SFT}}+\lambda\mathcal{L}_{\text{con}}$ with weight $\lambda>0$. Our
ablation finds that this term does not help (\Cref{sec:exp:abl}).

\subsection{Training Pipeline}
\label{sec:method:train}

\paragraph{Initialization.} By default the new creator rows are initialized randomly,
$\mathbf{w}_{\phi(a)}\sim\mathcal{N}(0,\sigma_0^2\mathbf{I})$ with standard deviation
$\sigma_0=0.02$. Optionally, they can be warm-started from the GNN embedding through a learned
linear projection $\mathbf{P}$, $\mathbf{w}_{\phi(a)}=\mathbf{P}\,E(a)$; unlike the SID codebook, this
serves only as an initialization and the embedding remains fully trainable. Creators added later
receive new rows without a codebook rebuild.

\paragraph{Stage 1: creator-token warm-up.} With the backbone frozen, we train only the new creator
rows on creator descriptions and co-engagement sequences, so that the tokens are grounded before
sequential training.

\paragraph{Stage 2: sequential SFT.} We adapt the backbone with Low-Rank Adaptation
(LoRA)~\citep{hu2022lora} and train the creator rows jointly with $\mathcal{L}_{\text{SFT}}$ on
(profile, history)$\,\to\,$next-creator examples.

\paragraph{Stage 3: post-training (optional).} Because the creator tokens are part of $\theta$, the
policy $\pi_\theta(a\mid z)=P_\theta(a\mid z)$ is differentiable over the true action space
$\mathcal{A}$, so it can be aligned from engagement with a value-weighted reward
\begin{equation}
r(a)=\sum_{\tau}v(\tau)\,\mathbb{1}\big[(a,\tau)\in\mathcal{F}_u\big],
\end{equation}
where $\mathcal{F}_u$ is the set of future (creator, engagement-type) pairs of user $u$ and the sum
runs over engagement types $\tau$. We evaluate Direct Preference Optimization
(DPO)~\citep{rafailov2023direct}, Group Relative Policy Optimization
(GRPO)~\citep{shao2024deepseekmath}, and Self-Distillation Policy Optimization
(SDPO)~\citep{hubotter2026reinforcement} from the SFT model in \Cref{app:pt}; none improves on it,
so the served model is the Stage-2 SFT model.

%% sections/related_work.tex
\section{Related Work}
\label{sec:related}

\paragraph{Generative recommendation.} Generative
recommendation~\citep{yun2026decoding,rajput2023recommender} has become a compelling alternative to
multi-stage filtering pipelines. It builds on a long line of sequential recommenders that model user
histories with recurrent, self-attentive, bidirectional, and self-supervised
encoders~\citep{hidasi2016session,kang2018self,sun2019bert4rec,zhou2020s3rec} and, more recently,
on large-scale generative transducers that recast ranking as sequence
generation~\citep{zhai2024actions}. A foundational design choice is the item identifier, which must
carry both semantic content and collaborative signal~\citep{rajput2023recommender,singh2024better}.
The dominant recipe tokenizes each item into a Semantic ID by discretizing a pretrained embedding
into a hierarchy of codes and training a language model to generate
them~\citep{rajput2023recommender,hou2023learning,jin2024language,zheng2024adapting,hou2025generating}.
MusicRec~\citep{Zhao2026MusicRecMS} and FusID~\citep{kim2026fusid} further fuse textual and acoustic
signals into the embedding before quantization~\citep{zhu2026general}. To make these continuous
embeddings usable by a large language model (LLM), vector-quantization
techniques~\citep{oord2017neural,razavi2019generating,defossez2023high} such as the residual
quantized variational autoencoder (RQ-VAE)~\citep{lee2022autoregressive} and residual quantization
with k-means (RQ-KMeans)~\citep{douze2025faiss} map each embedding to a tuple of codebook indices.
All of these methods place a many-to-one quantizer, fixed while the language model trains, between
the item and the language model; ACT removes it.

\paragraph{Preference optimization.} Reinforcement learning (RL) is the standard tool for
post-training alignment of
LLMs~\citep{guo2025deepseek,team2025kimi,olmo2025olmo,jaech2024openai,lambert2024tulu}. Methods
that learn from verifiable rewards, such as GRPO~\citep{shao2024deepseekmath}, optimize policy
gradients on outcome-level rewards, but sparse scalar feedback complicates credit assignment, and
process reward models~\citep{setlur2024rewarding} that supply step-level supervision require an
auxiliary model. In recommendation, DPO~\citep{rafailov2023direct} contrasts preferred and rejected
items at the item level, which does not decompose over multi-token SIDs; self-distillation
methods~\citep{zhao2026self,hubotter2026reinforcement,li2026rethinking,yang2026selfdistilled} instead
let the policy, conditioned on realized feedback, act as its own teacher. Because an ACT creator is
a single token, all three apply to ACT directly over the true action space; we find, however, that
none improves on SFT in our setting (\Cref{app:pt}).

%% sections/experiments.tex
\section{Experiments}
\label{sec:exp}

We compare ACT with the SID pipeline on production unconnected-creator data and on the public
KuaiRand dataset. We ask whether ACT improves creator-level recall (\Cref{sec:exp:main}), why
(\Cref{sec:exp:why}), how vocabulary size trades conditional recall against coverage
(\Cref{sec:exp:coverage}), whether the result transfers to public data and when it holds
(\Cref{sec:exp:public}), what each token layer costs to serve (\Cref{sec:exp:cost}), which design
choices matter (\Cref{sec:exp:abl}), and whether the offline gain carries over online
(\Cref{sec:exp:online}).

\subsection{Experimental Setup}
\label{sec:exp:setup}

\paragraph{Production data.} Next-creator examples come from engagements with suggested,
non-followed creators (opens, likes, shares, follows, and dwell), with auxiliary cross-surface
engagement as additional context and each history truncated to the most recent engaged creators. We
split by day so that no test interaction is seen in training, and the target is the next
unconnected creator on any surface. We evaluate three nested candidate sets by decreasing follower
threshold: \emph{top} ($N_A{=}Y$), \emph{mid} ($N_A{=}10Y$), and \emph{broad} ($N_A{=}30Y$), so a
larger vocabulary adds coverage rather than replacing creators. Vocabulary and codebook sizes are
reported in relative units: $Y$ is the smallest vocabulary and $X$ the base per-level codebook size,
so the SID codebooks are $X^3$, $(2X)^3$, and $(4X)^3$. The $10Y$ setting trains on ${\sim}10$M
engagement events and tests on ${\sim}1$M; the other settings use a test set of the same scale.

\paragraph{Public data.} To test transfer, we use KuaiRand-27K~\citep{gao2022kuairand}: the task is
to predict the next clicked creator from a user's 20 most recent creators. We restrict the
candidate set to the top-50k creators by training clicks ($90.4\%$ of training clicks), giving
27,065 users, 16.52M training and 100k test examples. Both token schemes use identical samples in
identical order and differ only in the creator token; both are trained with Stage~2 only (no
warm-up) for one epoch at learning rate $10^{-4}$. We add two heuristic baselines, \emph{global
popularity} (the most-clicked creators in training) and \emph{recent history} (the user's most
recent creators). To probe how much data ACT needs, we also train ACT on the smaller KuaiRand-1K,
both on all of its creators and on its top creators.

\paragraph{SID baseline.} The SID baseline follows OneRec~\citep{onerec2025}: a frozen GNN embedding
quantized by RQ-KMeans ($L{=}3$, $M{=}X$ unless stated) on the same creators, generated by the same
backbone, and decoded by codebook join and popularity rerank. Only the token layer differs from
ACT.

\paragraph{Metrics.} Because the served unit is a creator, we score both methods by
\emph{creator-level} Recall@$K$ ($K\in\{1,5,10,15,50,200\}$), i.e., whether the true creator is in
the top-$K$, and by normalized discounted cumulative gain (NDCG@200) weighted by the engagement
value $v(\tau)$. For ACT a token \emph{is} a creator. For SID we expand each predicted SID into its
member creators, ranked by beam order and then by follower count within a SID. We also report SID
\emph{cluster-level} recall, which counts a hit whenever the predicted SID equals the target's SID;
it overstates delivered recall because one SID stands for $\nu(s)$ creators, so only creator-level
recall is comparable to ACT.

\paragraph{Implementation.} Stage~1 trains the creator rows for one epoch at learning rate
$10^{-4}$; Stage~2 applies LoRA (rank $r{=}64$, scaling $\alpha{=}128$, dropout $0.05$) with the
creator rows trainable, at learning rate $2\times10^{-5}$ with a cosine schedule. Training runs in
bf16 with fully sharded data parallelism (FSDP) on $4$ hosts $\times$ $8$ graphics processing units
(GPUs); inference uses vLLM over the expanded vocabulary with beam width $15$ for $K{\le}15$ and
$200$ for the deeper cutoffs.

\subsection{Main Results}
\label{sec:exp:main}

\begin{figure}[htbp]
  \centering
  \includegraphics[width=\textwidth]{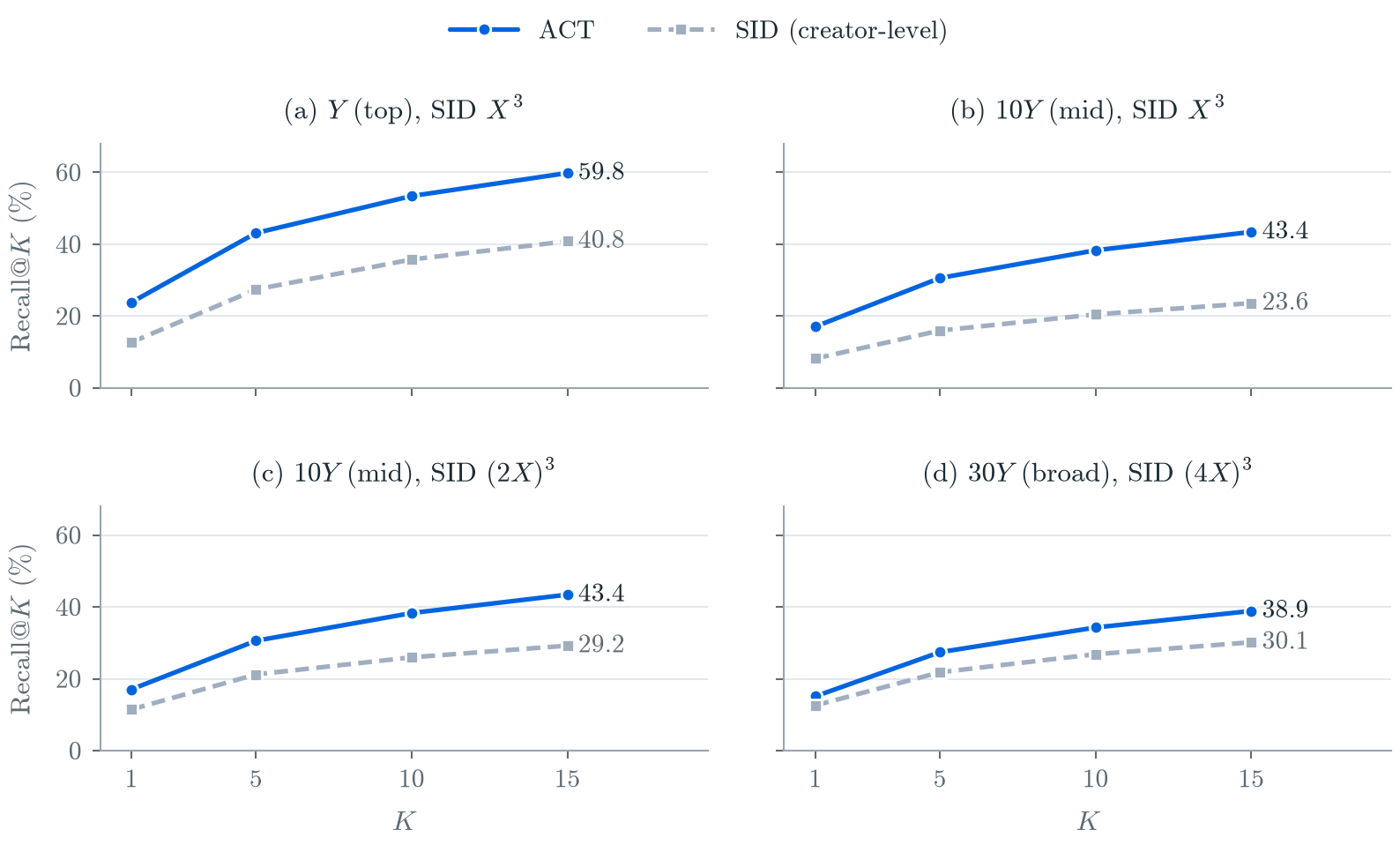}
  \caption{Creator-level Recall@$K$ of ACT and SID on production data at each vocabulary scale and
  SID codebook size. ACT leads at every $K$ in every setting.}
  \label{fig:recall_prod}
\end{figure}

ACT outperforms the SID baseline on every reported production metric at the $10Y$ scale with the
base $X^3$ codebook. Creator-level Recall@15 rises
from $23.58\%$ to $43.41\%$, Recall@50 from $34.33\%$ to $60.27\%$, and Recall@200 from $47.62\%$
to $79.30\%$. The corresponding absolute gains are $19.83$, $25.94$, and $31.68$ percentage
points: ACT's lead is present for short candidate lists and remains substantial for deeper ones.
NDCG@200 also increases from $0.187$ to $0.343$, a relative gain of $83.4\%$, indicating that the
recall improvement is accompanied by better performance under the rank-sensitive,
engagement-weighted metric. The advantage holds across the vocabulary and codebook settings in
\Cref{fig:recall_prod}, where ACT leads at every plotted cutoff ($K\in\{1,5,10,15\}$) in all four
production settings. With the SID codebook
fixed at $X^3$, ACT achieves Recall@15 of $59.80\%$ versus $40.81\%$ for SID at $Y$, and $43.41\%$
versus $23.58\%$ at $10Y$ (\Cref{tab:scale}). Both methods have lower conditional recall in the
larger vocabulary, but the absolute advantage of ACT stays close to $19$--$20$ points. Its relative
gain consequently widens from $46.5\%$ to $84.1\%$. Enlarging the SID codebook improves the
baseline but does not close this gap: at $10Y$, increasing the per-level size from $X$ to $2X$
expands the nominal code space eightfold and raises SID Recall@15 from $23.58\%$ to $29.21\%$.
ACT still reaches $43.41\%$, retaining a $14.20$-point advantage. At the broadest $30Y$ tier,
even the $(4X)^3$ codebook, with $64$ times the nominal capacity of $X^3$, yields $30.15\%$
Recall@15 against ACT's $38.87\%$. ACT also leads at every smaller plotted cutoff in this
setting, so its advantage persists beyond the smallest, most popular creator tier and against
codebooks with substantially more capacity. \Cref{sec:exp:why} examines the collision losses
behind this pattern, and \Cref{app:codebook} details the broadest-scale comparison and its
differing embedding coverage. All these recalls are scored at the creator level, after expanding
SID predictions into ranked creators, and thus measure retrieval of the actual target rather than
its semantic cluster. Since recall is conditional on the target belonging to each vocabulary,
the lower values at larger tiers do not by themselves establish which catalog size is best for
all engagement; \Cref{sec:exp:coverage} evaluates that tradeoff by combining conditional recall
with engagement coverage.

\FloatBarrier
\subsection{Why Atomic Creator Tokens Beat Semantic IDs}
\label{sec:exp:why}

SID predictions must identify a creator after identifying its semantic code, and collisions
introduce a loss at this second step. Creator-level recall factorizes as
\begin{equation}
R^{\text{SID}}_{\text{creator}}=R^{\text{SID}}_{\text{cluster}}\times\rho,\qquad\rho\in(0,1],
\label{eq:tax}
\end{equation}

Here, $R^{\text{SID}}_{\text{creator}}$ is SID's creator-level Recall@15, which counts a hit only
when the true creator appears in the final top-15 list. $R^{\text{SID}}_{\text{cluster}}$ is its
cluster-level Recall@15, which counts a hit when the target's SID is retrieved.
$\rho=R^{\text{SID}}_{\text{creator}}/R^{\text{SID}}_{\text{cluster}}$ is the resolution efficiency:
the fraction of cluster-level hits retained after expanding the predicted SIDs into creators and
applying the popularity tie-break. A value of $\rho=1$ means that this resolution step loses no
hits, while a smaller value indicates a larger loss.

\begin{figure}[!htb]
  \centering
  \includegraphics[width=\textwidth]{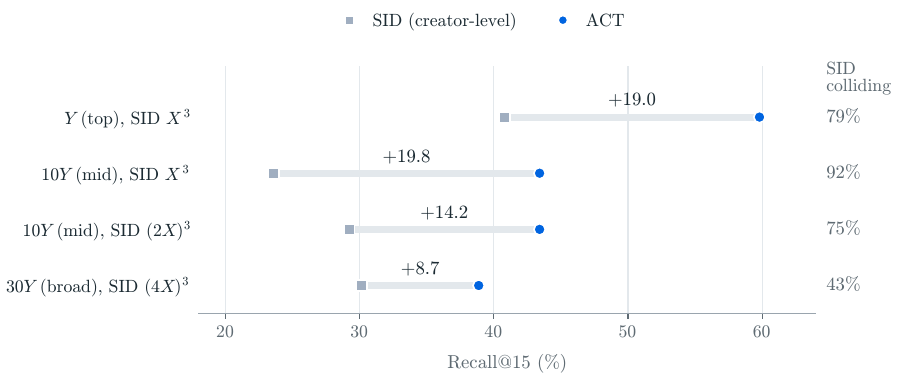}
  \caption{Creator-level Recall@15 of SID and ACT per setting, with the fraction of creators
  that share their SID. A larger codebook reduces collisions and narrows the gap, but the gap
  never closes.}
  \label{fig:collision_gap}
\end{figure}

Whenever $\nu>1$ and the tie-break is not user-specific, $\rho<1$, roughly
$\rho\lesssim1/\bar\nu$ for mean
collision count $\bar\nu$; ACT has $\nu\equiv1$ and hence $\rho\equiv1$. Thus, even with identical
cluster-level accuracy, SID loses a $(1-\rho)$ fraction of its hits, the \emph{collision tax}. This
is also why SID cluster-level recall can exceed ACT's creator-level recall while SID is the weaker
system: hitting the code does not identify the creator. The measured recalls in
\Cref{tab:scale} show that $\rho$ falls from $Y$ to $10Y$ as $\bar\nu$ rises under a fixed $X^3$
codebook. SID
is therefore doubly penalized at scale: the code is harder to predict \emph{and} the tax is
heavier, whereas ACT pays only the first cost, which helps explain why its relative advantage
widens with the catalog. Increasing the codebook size partly recovers these lost hits. Holding
the creators, data, and decoding fixed and varying only the per-level size, $(2X)^3$ at $10Y$
substantially raises $\rho$ and SID recall, supporting collisions as an important source of the
gap, yet ACT still leads (\Cref{fig:collision_gap}). Pushed to $30Y$ creators with
a $64\times$ larger $(4X)^3$ codebook, collisions fall monotonically with codebook size but never
reach zero, because the frozen GNN embeddings cluster intrinsically (\Cref{app:codebook}). ACT keeps
a clear lead even though SID is scored only on the creators that have a GNN embedding, while ACT is
scored on all of them. Enlarging the codebook buys diminishing returns; only $\nu\equiv1$ removes the
tax outright. The learned representations provide a complementary view of the remaining gap.
The data processing inequality discussed in \Cref{sec:method:act} relates SID's fixed quantization
to information loss, while ACT learns its creator embeddings directly under the sequential
objective. Consistent with this distinction, the learned creator embeddings are largely decoupled
from the frozen GNN prior: global, local, and layout-level agreement measures indicate little
shared structure (\Cref{app:repr}). ACT thus learns a task-specific geometry rather than simply
re-deriving the collaborative geometry encoded by the SID quantizer.

\begin{table}[!htb]
  \centering
  \caption{Recall@15 and collision statistics as the vocabulary and the SID codebook grow.
  $\bar\nu$: mean creators per SID; $\rho=R_{\text{creator}}/R_{\text{cluster}}$ for SID; tax
  $=R_{\text{cluster}}-R_{\text{creator}}$ (points); $\Delta$: gain of ACT over SID. ACT has
  $\bar\nu=\rho=1$ and no tax at every scale.}
  \label{tab:scale}
  \begin{tabular}{ll ccc ccc}
    \toprule
    Vocabulary & SID codebook & $\bar\nu$ & $\rho$ & Tax & SID & ACT & $\Delta$ \\
    \midrule
    $Y$ (top)      & $X^3$     & 2.35 & 0.61 & 25.9 & 40.81 & \textbf{59.80} & $+18.99$ \\
    $10Y$ (mid)    & $X^3$     & 3.79 & 0.47 & 26.8 & 23.58 & \textbf{43.41} & $+19.83$ \\
    $10Y$ (mid)    & $(2X)^3$  & 1.75 & 0.64 & 16.7 & 29.21 & \textbf{43.41} & $+14.20$ \\
    $30Y$ (broad)  & $(4X)^3$  & 1.37 & 0.74 & 10.4 & 30.15 & \textbf{38.87} & $+8.72$ \\
    \bottomrule
  \end{tabular}
\end{table}

\FloatBarrier
\subsection{Vocabulary Size and Coverage}
\label{sec:exp:coverage}

\begin{figure}[htbp]
  \centering
  \includegraphics[width=\textwidth]{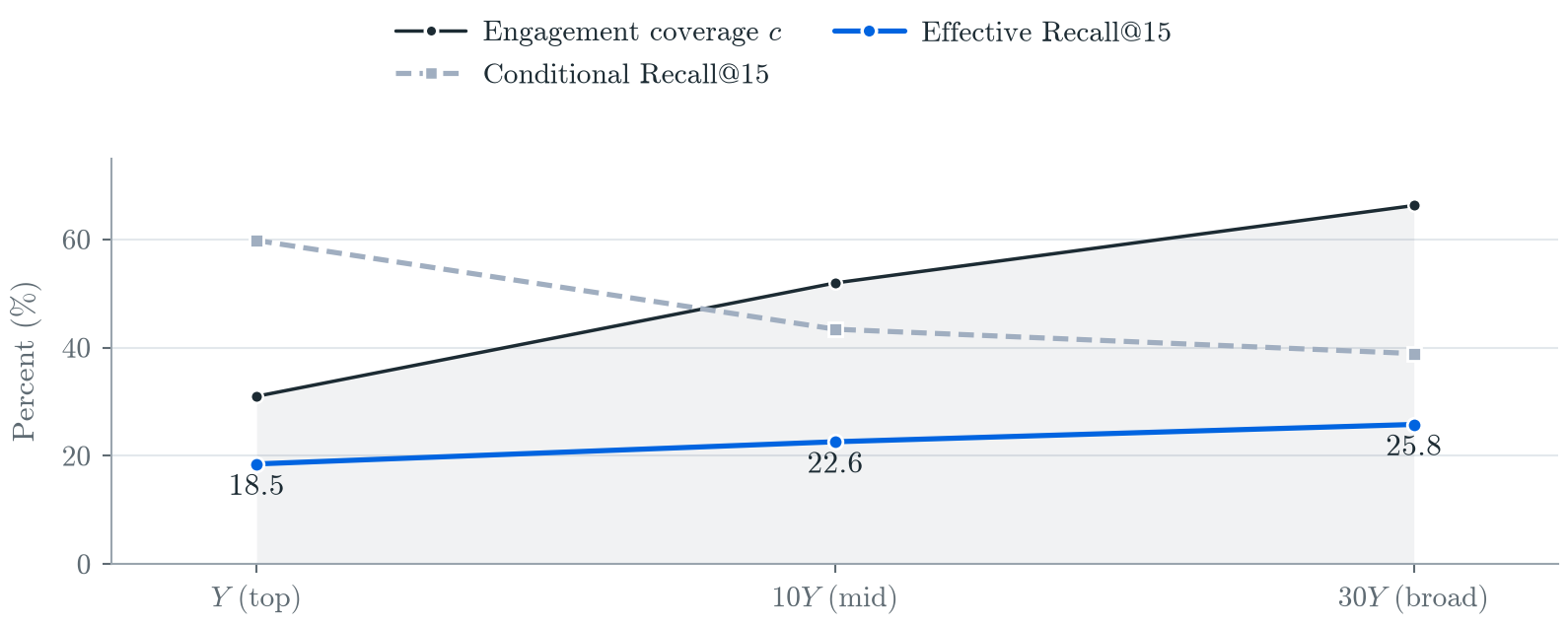}
  \caption{ACT across vocabulary tiers: conditional Recall@15 falls as the vocabulary grows, but
  engagement coverage $c$ rises faster, so effective Recall@15 ($c\times$conditional) rises
  monotonically.}
  \label{fig:coverage}
\end{figure}

A larger vocabulary makes each prediction harder but covers more of what users engage with. As the
ACT vocabulary grows from $Y$ to $30Y$, recall conditional on the target being in the vocabulary
falls, while the share of engagement the vocabulary covers rises faster, so the effective recall
over all engagement increases at every step (\Cref{fig:coverage}). Within the head regime, a larger
ACT vocabulary is therefore strictly better in delivered recall, which motivates scaling ACT up to
the point where per-token data runs out (\Cref{sec:exp:public}).

\FloatBarrier
\subsection{Public Data: KuaiRand}
\label{sec:exp:public}

\begin{figure}[htbp]
  \centering
  \begin{subfigure}[t]{0.47\textwidth}
    \centering
    \includegraphics[width=\linewidth]{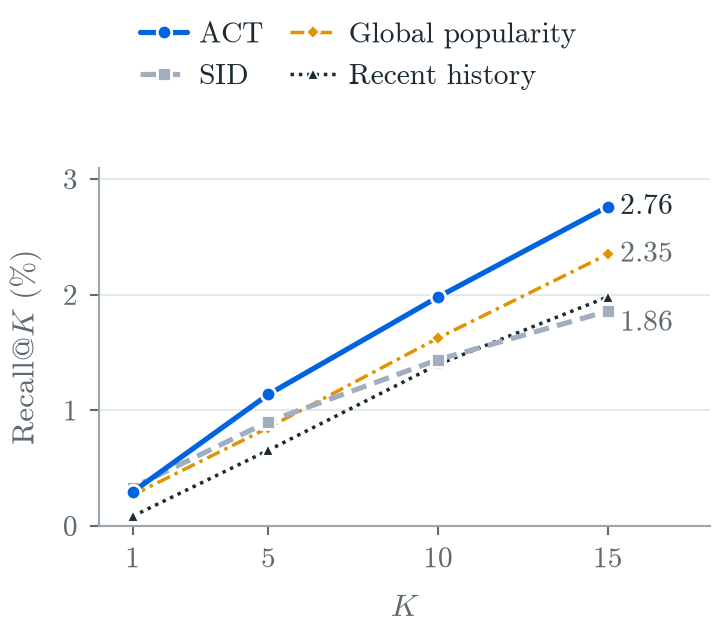}
    \caption{Creator-level Recall@$K$.}
    \label{fig:kuairand_recall}
  \end{subfigure}\hfill
  \begin{subfigure}[t]{0.47\textwidth}
    \centering
    \includegraphics[width=\linewidth]{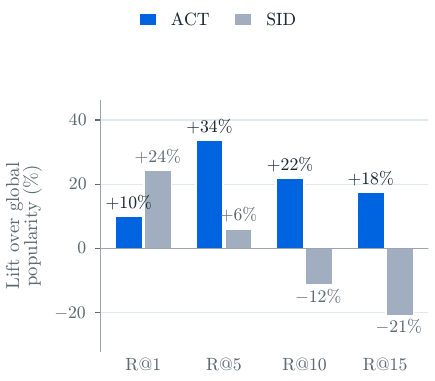}
    \caption{Relative lift over global popularity.}
    \label{fig:kuairand_lift}
  \end{subfigure}
  \caption{KuaiRand-27K (top-50k creators), with identical samples for ACT and SID. ACT is the only
  method that beats global popularity at every cutoff; SID falls below it from $K{=}10$. The SID
  codebook has $1.30$ creators per SID and a $36.2\%$ colliding fraction. Top-1 predictions cover
  12,260 distinct creators for ACT versus 4,399 distinct SIDs for SID.}
  \label{fig:kuairand}
\end{figure}

ACT's advantage over SID also appears on KuaiRand-27K, where both models use identical training
and test samples, the same sample order, and one epoch of Stage~2 training without vocabulary
warm-up. ACT leads SID at every evaluated cutoff from $K{=}5$ (\Cref{fig:kuairand}): Recall@5 is
$1.138\%$ versus $0.903\%$, and Recall@15 is $2.763\%$ versus $1.859\%$, corresponding to relative
gains of $26.0\%$ and $48.6\%$. The absolute gap also grows from $0.235$ to $0.904$ percentage
points, so the benefit extends beyond selecting a single creator. The two heuristic baselines
provide a useful check on what the models learn. Global popularity is competitive on this task,
reaching $2.351\%$ Recall@15, while recent history reaches $1.982\%$. ACT exceeds both and is the
only method above global popularity at every evaluated cutoff; its relative lift over popularity
is $33.7\%$ at $K{=}5$, $21.8\%$ at $K{=}10$, and $17.5\%$ at $K{=}15$. SID, in contrast, falls
below global popularity from $K{=}10$ and also below recent history at $K{=}15$. Its only advantage
over ACT is at $K{=}1$, where recall is $0.332\%$ versus $0.294\%$. This small top-1 lead is
consistent with the popularity prior introduced by returning the most popular creator within
each predicted SID, but it does not persist as the candidate list grows. The public-data codebook
has $1.30$ creators per SID and a $36.2\%$ colliding fraction, close to the collision profile of
the production $(4X)^3$ codebook. The result therefore holds with a substantially less colliding
quantizer than the base production codebook. The learned ACT vocabulary is also used broadly:
top-1 predictions span 12,260 distinct creators, compared with 4,399 distinct codes predicted by
SID on the same test set. However, the smaller KuaiRand-1K experiments show that atomic tokens do
not automatically outperform simple retrieval heuristics. With all creators included, ACT
achieves $0.75\%$ Recall@15 against $1.76\%$ for the best heuristic; restricting the vocabulary
to top creators raises ACT recall to $1.86\%$, still below the heuristic's $3.78\%$. These settings
provide about $5$ and $34$ training samples per creator token on average, compared with about
$331$ in the KuaiRand-27K top-50k setting. The larger dataset also supplies more total training
data, so this comparison does not isolate a universal samples-per-token threshold. It does show
that the public-data gain depends on the available supervision, rather than following from the
one-token representation alone. Together, these results support the routed design of
\Cref{sec:intro}: use ACT for well-observed creators in the head and conventional retrieval where
creator-level training data remains sparse.

\FloatBarrier
\subsection{Inference Cost}
\label{sec:exp:cost}

\begin{figure}[htbp]
  \centering
  \includegraphics[width=\textwidth]{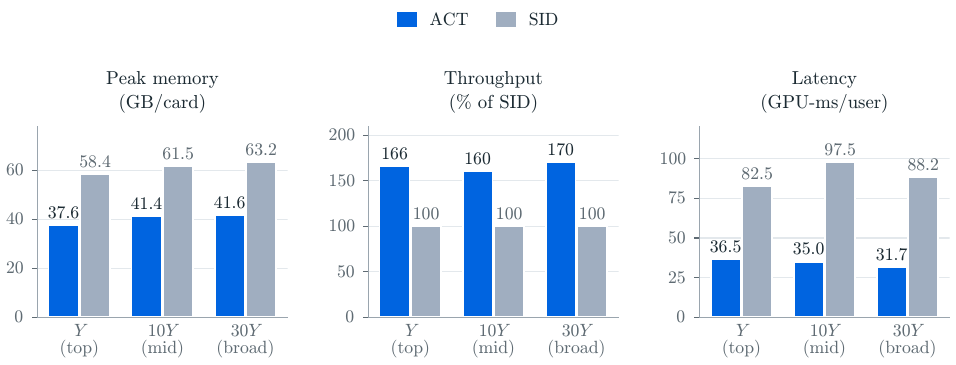}
  \caption{Inference cost of ACT and SID across vocabulary tiers ($32\times$A100-80GB, beam width
  15, ${\sim}1$M-user test). Throughput is concurrent requests at fixed memory, relative to SID.}
  \label{fig:cost}
\end{figure}

ACT is also cheaper to serve at every scale (\Cref{fig:cost}). Under the same hardware
($32\times$A100-80GB), beam width $15$, and ${\sim}1$M-user test set, ACT needs less GPU memory per
card, serves more concurrent requests at a fixed memory budget, and has lower per-user latency.
The driver is one token versus five: SID emits a five-token block
($\langle$bos$\rangle\,s_1s_2s_3\,\langle$eos$\rangle$), which multiplies the decode steps and the
key--value (KV) cache held by every beam, whereas the only extra cost of ACT is a wider softmax over
the expanded vocabulary, a single bandwidth-friendly matrix multiplication. The token-layer choice
is therefore a Pareto improvement in both accuracy and serving cost.

\FloatBarrier
\subsection{Ablation Study}
\label{sec:exp:abl}

GNN warm-starting yields modest recall gains over random initialization (\Cref{tab:abl}). Adding
the in-batch contrastive auxiliary loss to the warm-started variant reduces recall at every
reported cutoff. We use the simpler model with randomly initialized, trainable creator embeddings
for serving. Although ACT makes on-policy alignment feasible, none of DPO, GRPO, or SDPO
improves on the SFT model (\Cref{app:pt}), which keeps reinforcement learning off the serving
critical path.

\begin{table}[htbp]
  \centering
  \caption{Ablations of ACT design choices (SFT, $10Y$; creator-level Recall@$K$ with beam
  width 200; best per column in \textbf{bold}).}
  \label{tab:abl}
  \begin{tabular}{l ccc}
    \toprule
    Variant & R@15 & R@50 & R@200 \\
    \midrule
    Random init, trainable          & 43.91 & 60.96 & 80.01 \\
    GNN warm-start, trainable       & \textbf{44.66} & \textbf{61.62} & \textbf{80.44} \\
    \quad + contrastive loss        & 43.39 & 59.54 & 77.22 \\
    \bottomrule
  \end{tabular}
\end{table}

\FloatBarrier
\subsection{Online Test}
\label{sec:exp:online}

We evaluate ACT in a live four-arm A/B test against the SID-enhanced production arm (prod+SID) and
the classic production ranker (prod). The ACT arm has a positive directional lift for every
reported metric against both controls (\Cref{tab:ab}). The largest change is in product engagement:
$+0.29\%$ over prod+SID and $+0.66\%$ over prod, with the latter approaching statistical
significance ($p\approx0.05$). The other changes are small and are best interpreted as directional
evidence that the offline recall gain transfers online.

\begin{table}[htbp]
  \centering
  \caption{Online relative change (\%) of ACT against the SID-enhanced production arm (prod+SID)
  and the classic production ranker (prod). The product-engagement change against prod approaches
  statistical significance ($p\approx0.05$).}
  \label{tab:ab}
  \begin{tabular}{l cc}
    \toprule
    Metric & vs.\ prod+SID & vs.\ prod \\
    \midrule
    Overall consumption metric 1 & $+0.11$ & $+0.05$ \\
    Overall consumption metric 2 & $+0.01$ & $+0.01$ \\
    Product consumption metric   & $+0.02$ & $+0.03$ \\
    Product engagement metric    & $+0.29$ & $+0.66$ \\
    \bottomrule
  \end{tabular}
\end{table}

%% sections/conclusion.tex
\section{Conclusion}
\label{sec:conclusion}

In cold-start unconnected creator recommendation, the binding constraint is the SID layer, not the
language model: a surjective creator-to-SID map imposes a collision-limited recall ceiling, and a
frozen quantizer blocks end-to-end learning. ACT removes this layer with a bijective per-creator
vocabulary trained end-to-end. Under a controlled comparison, ACT improves creator-level Recall@15
by $+19.8$ points ($1.84\times$) over a matched SID baseline, keeps its lead as the catalog and the
codebook grow, transfers to the public KuaiRand dataset, and is cheaper to serve. Its premise is
that every creator token sees enough training data, so ACT is best deployed as the head of a
routed system, with the long tail served by conventional retrieval. Multi-interest decoding,
hierarchical tokens for latency-critical settings, and on-policy alignment for objectives that SFT
cannot express are natural next steps.

%% ---------------------------------------------------------------------------
\clearpage
\newpage
\bibliographystyle{assets/plainnat}
\bibliography{paper}

%% ---------------------------------------------------------------------------
\clearpage
\newpage
\beginappendix

%% sections/appendix.tex
\section{Largest-Scale Codebook Stress Test}
\label{app:codebook}

\Cref{tab:codebook} details the stress test of \Cref{sec:exp:why}: a $30Y$-creator catalog with the
SID codebook grown from $X^3$ to $(4X)^3$. Collisions shrink monotonically with codebook size but
never vanish: even the largest codebook leaves a large share of creators sharing a code, because the
frozen GNN embeddings cluster intrinsically. Against the strongest SID, ACT still leads at every
cutoff, and the comparison favors SID, which is scored only on the ${\sim}76\%$ of the $30Y$
creators that have a GNN embedding, whereas ACT is scored on all of them.

\begin{table}[h]
  \centering
  \caption{Codebook stress test ($30Y$ catalog, $L{=}3$). Top: collisions against codebook size.
  Bottom: creator-level Recall@$K$ of the largest SID codebook and ACT (beam width 15, ${\sim}1$M
  test examples).}
  \label{tab:codebook}
  \begin{tabular}{l cccc}
    \toprule
    \multicolumn{5}{l}{\emph{Collisions against codebook size}} \\
    \midrule
    Codebook & Reachable codes & Used / reachable & $\bar\nu$ & Colliding creators \\
    \midrule
    SID $X^3$       & $X^3$     & 88.7\% & 8.06 & 98.4\% \\
    SID $(2X)^3$    & $(2X)^3$  & 43.5\% & 2.06 & 73.2\% \\
    SID $(4X)^3$    & $(4X)^3$  & 8.2\%  & 1.37 & 43.3\% \\
    ACT (bijective) & --        & --     & 1.00 & 0\%    \\
    \midrule
    \multicolumn{5}{l}{\emph{Creator-level Recall@$K$ (\%): largest SID codebook against ACT}} \\
    \midrule
    Model & R@1 & R@5 & R@10 & R@15 \\
    \midrule
    SID $(4X)^3$ & 12.57 & 21.80 & 26.83 & 30.15 \\
    ACT          & \textbf{15.10} & \textbf{27.39} & \textbf{34.27} & \textbf{38.87} \\
    \bottomrule
  \end{tabular}
\end{table}

\section{Non-Generative Retrieval Baselines}
\label{app:baselines}

The main text compares ACT with a matched SID model, the representation the generative recipe
directly replaces. For completeness, we also place ACT against the non-generative retrieval it
would sit next to in a routed system: the deployed creator-to-creator (C2C) retrieval stack. We
report two of its live sources (A and B, identities masked) and the reciprocal-rank-fusion (RRF)
ensemble C that unions them and is the one actually served. All methods are scored on the same
${\sim}1$M-user test set by creator-level Recall@15, conditional on the target falling inside the
vocabulary at each scale; the generative models decode with beam width 15, and the retrieval sources
return their native ranked lists.

\Cref{tab:baselines} shows a clean ladder at every scale. The C2C sources retrieve few in-vocabulary
candidates for these unconnected creators, so their recall is coverage-limited by construction. They
are general-purpose retrievers tuned for the broad catalog rather than the popularity-scoped head
that ACT targets, so the comparison is directional rather than matched. Still, the matched SID model
clearly outperforms the best retriever, and ACT leads throughout, which supports reserving
conventional retrieval for the long tail.

\begin{table}[h]
  \centering
  \caption{ACT against non-generative retrieval baselines: creator-level Recall@15 (\%) on the
  in-vocabulary population, ${\sim}1$M test users; best per column in \textbf{bold}.}
  \label{tab:baselines}
  \begin{tabular}{l ccc}
    \toprule
    Method & $Y$ (top) & $10Y$ (mid) & $30Y$ (broad) \\
    \midrule
    C2C source A        & 6.70 & 5.58 & 5.04 \\
    C2C source B        & 4.12 & 5.30 & 5.68 \\
    C2C ensemble C      & 9.45 & 9.53 & 9.41 \\
    \midrule
    SID (creator-level) & 40.81 & 23.58 & 30.15 \\
    ACT                 & \textbf{59.80} & \textbf{43.41} & \textbf{38.87} \\
    \bottomrule
  \end{tabular}
\end{table}

\section{Post-Training}
\label{app:pt}

Because ACT's creator embeddings are policy parameters, the policy $\pi_\theta(a\mid z)$ is
differentiable over the true action space, so we ask whether on-policy alignment also helps. We
apply three alignment methods to a shared SFT reference policy
$\pi_{\text{ref}}=\pi_{\theta_{\text{SFT}}}$, with the value-weighted reward $r(a)$ of
\Cref{sec:method:train}.

\paragraph{DPO}~\citep{rafailov2023direct} uses item-level pairs of a preferred creator $a^+$ and a
rejected creator $a^-$ and minimizes
\begin{equation}
\mathcal{L}_{\text{DPO}}=-\mathbb{E}_{(z,a^+,a^-)}\!\left[\log\sigma\!\Big(\beta
\log\tfrac{\pi_\theta(a^+\mid z)}{\pi_{\text{ref}}(a^+\mid z)}-\beta
\log\tfrac{\pi_\theta(a^-\mid z)}{\pi_{\text{ref}}(a^-\mid z)}\Big)\right],
\end{equation}
where $\sigma$ is the logistic function, $\beta>0$ controls the strength of the implicit KL
constraint to $\pi_{\text{ref}}$, and the expectation is over contexts with a preference pair.

\paragraph{GRPO}~\citep{shao2024deepseekmath} samples a group of $G$ on-policy creators
$\{a_i\}_{i=1}^{G}\sim\pi_{\theta_{\text{old}}}(\cdot\mid z)$ from the previous policy
$\pi_{\theta_{\text{old}}}$, scores each with reward $r_i=r(a_i)$, and forms group-normalized
advantages $\hat{A}_i=(r_i-\mathrm{mean}(\mathbf{r}))/\mathrm{std}(\mathbf{r})$ over the group
rewards $\mathbf{r}=(r_1,\dots,r_G)$, updating $\theta$ with the clipped policy-gradient surrogate and
a KL penalty to $\pi_{\text{ref}}$. Because an ACT creator is a single token, the diversity of a
group is exactly the entropy of $\pi_\theta(\cdot\mid z)$: a low-entropy SFT model yields
near-duplicate rollouts, $\mathrm{std}(\mathbf{r})\to0$, and the advantages, hence the gradient,
vanish.

\paragraph{SDPO}~\citep{zhao2026self,hubotter2026reinforcement,li2026rethinking,yang2026selfdistilled}
needs no external reward model: it conditions the policy on its own realized engagement feedback $f$
to form a self-teacher $\pi_\theta(\cdot\mid z,f)$ that supplies dense targets, and trains the
student $\pi_\theta(\cdot\mid z)$ to match it. Its signal is proportional to the teacher--student
gap, which collapses when conditioning on $f$ barely changes the distribution.

\paragraph{Results.} No recipe beats SFT (\Cref{tab:pt}). DPO is compared against the main SFT
model; GRPO and SDPO run through the on-policy evaluation flow and are compared against its own
reference. The best DPO run only matches SFT, and GRPO and SDPO fall below their reference. The
cause is structural rather than a tuning artifact: a single-token, low-entropy policy over a
$10Y$-way vocabulary yields near-duplicate rollouts and a sparse, low-variance reward, whereas ACT
already casts recommendation as dense, exactly labeled token prediction that SFT optimizes
directly. For a production system this outcome is welcome: reinforcement learning need not sit on
the serving critical path.

\begin{table}[h]
  \centering
  \caption{Post-training the ACT SFT model with DPO, GRPO, and SDPO ($10Y$; creator-level
  Recall@$K$ in \%, beam width 15; best configuration per method). $\Delta$@15 is relative to the
  applicable reference.}
  \label{tab:pt}
  \begin{tabular}{l ccccc}
    \toprule
    Method & R@1 & R@5 & R@10 & R@15 & $\Delta$@15 \\
    \midrule
    SFT (reference)         & 17.04 & 30.57 & 38.29 & \textbf{43.41} & -- \\
    \quad + DPO (best)      & 17.07 & 30.49 & 38.20 & 43.26 & $-0.15$ \\
    \midrule
    RL-flow reference       & 13.88 & 23.83 & 29.56 & 33.47 & -- \\
    \quad + GRPO (best)     & 14.44 & 23.37 & 28.61 & 32.20 & $-1.27$ \\
    \quad + SDPO (best)     & 11.19 & 18.98 & 23.59 & 26.71 & $-6.77$ \\
    \bottomrule
  \end{tabular}
\end{table}

\section{Representation Analysis: ACT against the GNN Prior}
\label{app:repr}

\begin{figure}[h]
  \centering
  \includegraphics[width=\textwidth]{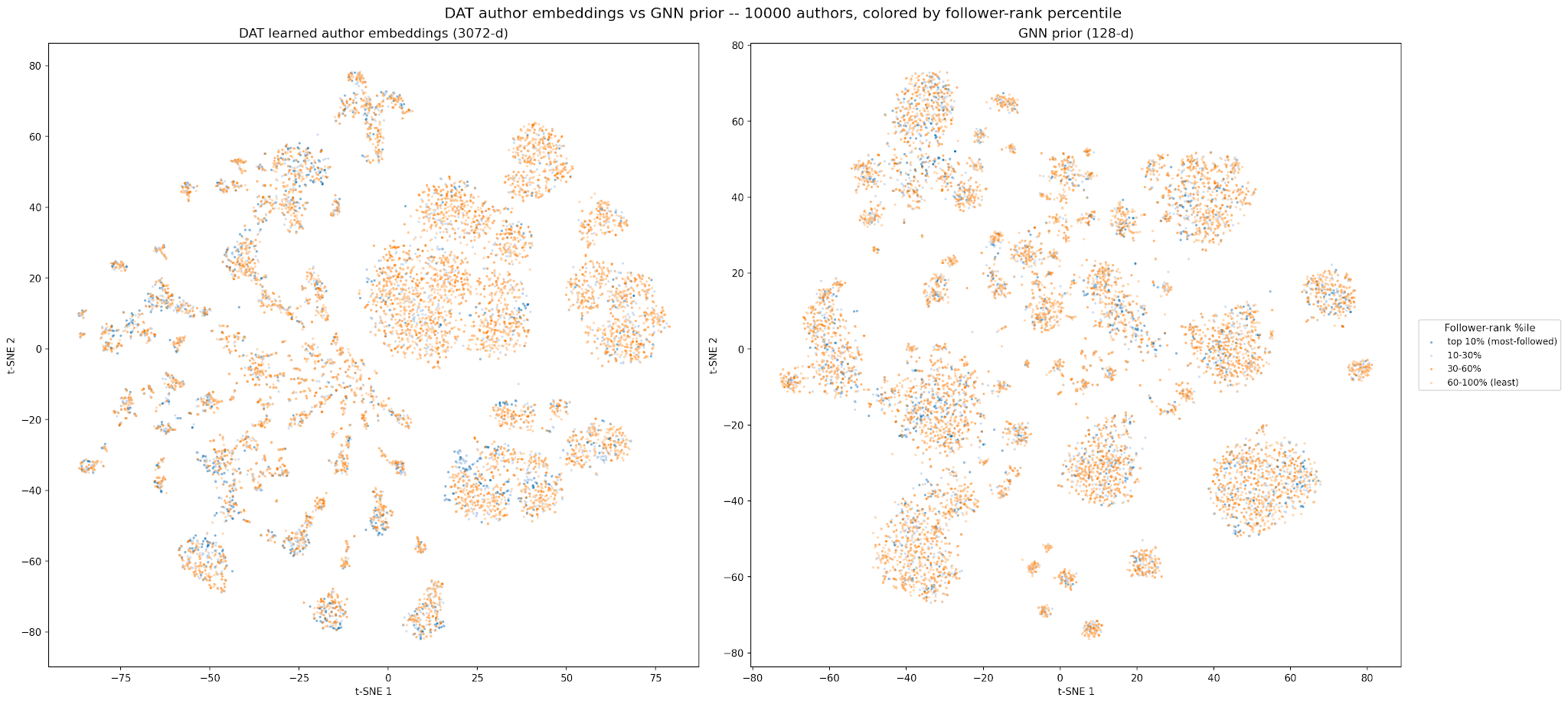}
  \caption{Independent t-SNE of the ACT creator embeddings (3072-d) and the production GNN prior
  (128-d) for the same 10,000 creators. The similar look is a t-SNE artifact; \Cref{tab:align}
  shows the two spaces are largely decoupled.}
  \label{fig:tsne}
\end{figure}

\Cref{sec:exp:why} argued that ACT learns a task-specific creator geometry rather than re-deriving
the frozen GNN prior. We test this by comparing the $3072$-dimensional creator-token rows of the
$10Y$ model with the $128$-dimensional production GNN embeddings of the same $10{,}000$ creators,
using independent t-SNE projections (\Cref{fig:tsne}) and three measures of structural agreement
(\Cref{tab:align}).

We use three complementary measures because no single one is decisive. Linear centered kernel
alignment (CKA) captures global alignment and is invariant to orthogonal transforms and isotropic
scaling, so the dimensionality gap does not bias it. The $k$-nearest-neighbor overlap (kNN@10)
probes local structure: the fraction of each creator's ten nearest GNN neighbors that stay among its
ten nearest in ACT space. Procrustes disparity, computed on the two t-SNE layouts after the optimal
rotation, reflection, and scaling, measures how far the layouts diverge once that freedom is
removed. All three would sit near their ``identical'' end if ACT merely re-derived the GNN prior.

Instead, all three show that the two representations are largely decoupled (\Cref{tab:align}).
Global agreement is close to zero; local overlap is well above chance, so a thin residue of
collaborative locality survives, yet most of each creator's GNN neighborhood is reorganized; and
after optimal alignment the same creator lands far apart in the two layouts. The two t-SNE panels
look superficially similar, a known t-SNE artifact; the quantitative measures, not the pictures,
carry the conclusion. The cause is architectural: the creator rows are randomly initialized and
learn their own geometry under the sequential objective, whereas an SID model is confined to the
geometry the frozen quantizer already encodes.

\begin{table}[h]
  \centering
  \caption{Structural agreement between the ACT creator embeddings and the frozen GNN prior
  ($n{=}10{,}000$ creators).}
  \label{tab:align}
  \begin{tabular}{l c l}
    \toprule
    Measure & Value & Reading \\
    \midrule
    Linear CKA                    & 0.10 & near-zero shared global structure (1 = identical) \\
    kNN@10 overlap                & 0.07 & ${\sim}70\times$ the random baseline of 0.001; 93\% of neighbors change \\
    Procrustes disparity (t-SNE)  & 0.887 & layouts nearly uncorrelated (1 = uncorrelated) \\
    \bottomrule
  \end{tabular}
\end{table}

\end{document}